\documentclass[a4paper, 10pt, conference]{ieeeconf}      % Use this line for a4
\usepackage{FG2024}
\usepackage{graphicx}
\usepackage{amsmath}
\usepackage{amssymb}
\usepackage{float}
\usepackage{multirow}
\usepackage{bm}
\usepackage{url}
\usepackage{subfigure}
\usepackage{colortbl}

\usepackage[noadjust]{cite}
\FGfinalcopy % *** Uncomment this line for the final submission

\IEEEoverridecommandlockouts                              % This command is only
\def\FGPaperID{0023} % *** Enter the FG2024 Paper ID here

\title{\LARGE \bf
Hierarchical Generative Network for Face Morphing Attacks
}
\author{\parbox{16cm}{\centering
    {\normalsize Zuyuan He, Zongyong Deng, Qiaoyun He and Qijun Zhao$^*$}\\
    {\small
    College of Computer Science, SiChuan University, Chengdu, China\\}}
	\thanks{$^*$Corresponding author}
    \thanks{This work is supported by the National Natural Science Foundation of China (No. 61773270, 62176170).}% <-this % stops a space
}

\usepackage{fancyhdr}
\begin{document}

\ifFGfinal
\thispagestyle{empty}
\pagestyle{empty}
\else
\author{Anonymous FG2024 submission\\ Paper ID \FGPaperID \\} % \FGPaperID
\pagestyle{plain}
\fi
\maketitle
\thispagestyle{fancy}

\begin{figure*}[ht]
	%		\vspace{-4mm}
	%\begin{center}
	\centering
	\setlength{\abovecaptionskip}{-0.04cm} %调整图片与标题距离
	\includegraphics[width=0.9\linewidth]{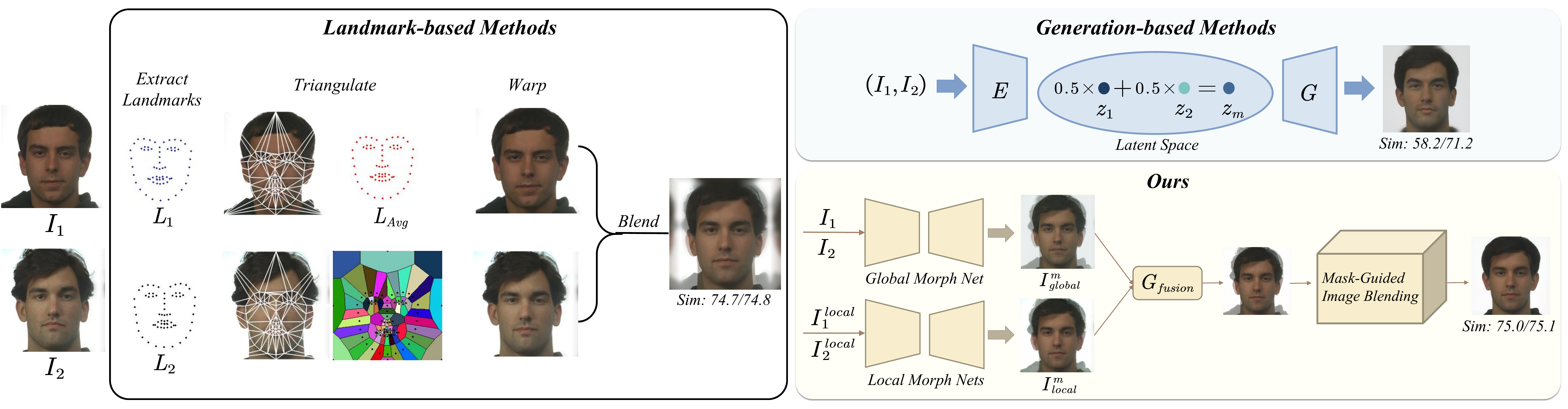} %[width=0.8\linewidth]
	\caption{Illustration of different face morphing methods. `\emph{Sim}' denotes the similarity between morphed and contributing images, which is computed by using ElasticFace~\cite{elasticface}.}
	\label{fig:insight}
	%	\vspace{-7mm}
	%\end{center}
\end{figure*}

%%%%%%%%%%%%%%%%%%%%%%%%%%%%%%%%%%%%%%%%%%%%%%%%%%%%%%%%%%%%%%%%%%%%%%%%%%%%%%%%
\begin{abstract}

Face morphing attacks circumvent face recognition systems (FRSs) by creating a morphed image that contains multiple identities. However, existing face morphing attack methods either sacrifice image quality or compromise the identity preservation capability. Consequently, these attacks fail to bypass FRSs verification well while still managing to deceive human observers. These methods typically rely on global information from contributing images, ignoring the detailed information from effective facial regions. To address the above issues, we propose a novel morphing attack method to improve the quality of morphed images and better preserve the contributing identities. Our proposed method leverages the hierarchical generative network to capture both local detailed and global consistency information. Additionally, a mask-guided image blending module is dedicated to removing artifacts from areas outside the face to improve the image's visual quality. The proposed attack method is compared to state-of-the-art methods on three public datasets in terms of FRSs' vulnerability, attack detectability, and image quality. The results show our method's potential threat of deceiving FRSs while being capable of passing multiple morphing attack detection (MAD) scenarios.

\end{abstract}

%%%%%%%%%%%%%%%%%%%%%%%%%%%%%%%%%%%%%%%%%%%%%%%%%%%%%%%%%%%%%%%%%%%%%%%%%%%%%%%%
\section{INTRODUCTION}

Face Recognition Systems (FRSs)~\cite{elasticface, arcface, facenet, curricularface} have been extensively deployed in critical security applications, such as Automatic Border Control (ABC) and financial services. However, FRSs can be vulnerable to various attacks like face forgery, and face morphing attacks~\cite{cao2022review,venkatesh2021face}. Where the face morphing attack is to create an image with multiple identities. 
%However, FRSs can be vulnerable to face morphing attacks that create one image with multiple identities~\cite{cao2022review}. 
Ferrara \emph{et al}.~\cite{matteo2014magic} demonstrated that a single face morphing attack image can successfully match with more than one person, disrupting the one-to-one mapping between original face images and individual identity. To better protect FRSs from morphing attacks, it is important to understand how to generate high quality morphing attack images, especially considering the recent rapid advances in generative models.
%Existing face morphing attack techniques manipulate facial landmarks and texture to create convincing morphs. Despite their impressive performance, producing high-quality morphed images that are visually flawless and can accurately resemble multiple individuals remains a difficult challenge.

In the literature, face morphing methods can be divided into two categories: landmark-based~\cite{Alpher08,Alpher09,raghavendra2017face} and generation-based~\cite{Alpher02,Alpher03,Alpher04,Alpher05,2023diffusion,2023mordiff}. Landmark-based morphing methods 
%generate morphed images by interpolating facial landmarks and blending textures. 
create face morphing attacks on the image level, usually by interpolating the contributors' facial landmarks to perform Delaunlay triangulation, warping the contributing images to align the interpolated average landmarks, and then alpha blending the two warped images to obtain the morphed image. Such methods are excellent at preserving the identity of the contributing images and pose a greater threat to FRSs. Nevertheless, artifacts like blurring and ghosting are created in areas around the eyes, nose, mouth, and hair due to insufficient landmarks detection during morphing. To alleviate artifacts, existing landmark-based morphing methods resort to additional post-processing steps. 
%If high-quality morphed images are to be obtained, artifacts need to be removed by additional post-processing steps. 
ReGenMorph~\cite{2021regenmorph} uses Generative Adversarial Networks (GAN) as a post-processing module to remove blending artifacts and gets visibly realistic morphed images. However, it greatly decreased the attack capability of the morphed images. Hence, while ensuring the attack capability of morphs, improving their visual quality further remains a challenge.

%Hence an end-to-end architecture of morphing methods is necessary. 

Unlike landmark-based methods, the generation-based methods are performed on the latent level. The contributing images are projected into the latent space, and their identities are interpolated to synthesize the morphed images. Such morphs exhibit improved visual quality, yet minor artifacts persist. However, these methods do not effectively preserve the identities of contributing images. Most of these methods rely on a single network to extract global information, such as shapes and textures. Local information that can better capture subtle facial features and variations is ignored, for instance, the color of the eyes or the shape of the nose and mouth~\cite{borghi2021automated,2020low}. 
%In~\cite{2020low}, it introduces a partial face manipulation-based morphing method that only focuses on two specific regions (eyes and nose). However, it is actually a post-processing improvement for the existing landmark-based method, not a GAN-based one. 
This motivates us to use GANs with different structures to extract both global and local information  from contributing images, facilitating the generation of morphs that are more threatening to FRSs.
%This motivates us to create a face morphing method that combines local and global information to enhance the quality of morphed images.

In this paper, we propose an end-to-end Hierarchical Generative network for Face Morphing (HGFM) attacks. Fig.~\ref{fig:insight} highlights the main difference between our method and existing ones. Traditional landmark-based methods suffer from solid artifacts outside the facial area when landmarks are missing, while generation-based methods sacrifice detailed biometrics and produce blurry morphs. In contrast, our method effectively extracts local detailed and global consistency information from facial images, resulting in morphs with better visual quality. The principal contributions of this paper are outlined as follows:
\begin{itemize}
	\item We devise a hierarchical structure comprising one global morph network and six local morph networks to extract local and global information for morphed face image generation.
	%	\item We design novel loss functions that constrain facial geometry and appearance in the generation process to improve the success rate of attacking FRSs.
	\item We design a loss function with five loss terms, including a novel combined identity loss (to improve the generalization ability of morphed images) and a geometry loss (to preserve the face geometric information of both contributors).
	\item We propose a mask-guided image blending module that effectively removes artifacts to improve the visual quality and preserve identity better. 
\end{itemize}
\
\indent We evaluate the proposed HGFM on three face datasets and compare them to publicly available state-of-the-art morphing methods. The results demonstrate that HGFM generates morphed images with favorable visual quality, better identity preservation, and a higher threat to FRSs.

\section{Related Work}
%\textbf{Face Morphing Attacks}.
\subsection{Face Morphing Attacks}
Existing face morphing attacks are performed either on image-level or on latent-level. Landmark-based morphing methods create face morphing attacks on image-level, and the morphed images are obtained by interpolating facial landmarks and blending the texture.
%Current methods for face morphing attacks are mainly categorized into landmark-based and GANs-based~\cite{Alpher02,Alpher03,Alpher04,Alpher05}, on image-level and representation-level, respectively.
%Landmark-based morphing methods create face morphing attacks on image-level, usually by interpolating the contributings' facial landmarks to perform Delaunlay Triangulation, warping the contributing images to align the interpolated average landmarks, and finally performing blending of the two wrapped images to obtain the morphed images. 
%Ferrera \etal~\cite{Alpher06} and Ramachandra \etal~\cite{Alpher07} have designed a series of variants based on this.
Among the open-source face morphing, GIMP/GAP, OpenCV~\cite{Alpher08}, and FaceMorpher~\cite{Alpher09} rely on landmarks. The morphs generated by these methods have apparent artifacts that appear unrealistic and easily detectable by human observers.
%During the morphing process, transferring landmarks and textures can lead to mismatched pixel positions that generate ghost-like images. These anomalies make the morphs appear unrealistic and easily detectable by human observers. Some manual post-processing methods can be used to remove artifacts. 
ReGenMorph~\cite{2021regenmorph} used an additional GAN to eliminate artifacts. In~\cite{2020low}, it introduced a partial face manipulation-based morphing method that only focuses on two specific regions (eyes and nose). However, this is tedious and reduces the performance of morphs.
Generation-based morphing methods typically use generation models like GAN and diffusion models to create morphed images by merging two facial images on the latent level. MorGAN, proposed by Damer \emph{et al}.~\cite{Alpher02}, was the first to use GAN to generate morphs with minor artifacts, which can not preserve identity well and are limited in low resolution ($64\times64$ pixels). Later, Damer \emph{et al}.~\cite{2019realistic} proposed an image enhancement solution to increase the quality and resolution of GAN-based morphs.
%that encodes the images used for morphing into latent space and interpolates the latent representations. Finally, the GAN generator receives the interpolated latent vector and creates a morphed image. Although the generated morphed image has minor artifacts, it cannot preserve identity and is limited in low resolution ($64\times64$ pixels). 
Inspired by MorGAN, Venkatesh \emph{et al}.~\cite{Alpher03} used StyleGAN~\cite{Alpher04} to generate realistic morphs with both high quality and high resolution ($1024\times1024$ pixels). Then, MIPGAN-I and MIPGAN-II~\cite{Alpher05} used the StyleGAN architecture to generate morphs with improved identity preservation by introducing a novel loss function. Recently, Blasingame~\cite{2023diffusion} and Damer \emph{et al.}~\cite{2023mordiff} investigated using diffusion models to create morphs. He \emph{et al.}~\cite{he2023optimal} produced high realistic morphs by optimizing morphing landmarks and using Graph Convolutional Networks (GCNs) to combine landmarks and appearance features.

%\noindent\textbf{Image Blending}. 
\subsection{Image Blending}
Image blending is widely used as a post-processing method to remove artifacts from morphed images. It involves taking a specific area from one image and seamlessly integrating it into a specific location in another image to create a composite image that appears natural. As described in~\cite{Alpher12}, Poisson image blending is a widely employed technique for achieving smooth gradient transitions in composite images. 
%Furthermore, in many cases, the colors of the target image tend to overpower the original source object, leading to a significant loss of source object content. 
Inspired by this work, we use a mask to identify the face region and separate it from the background region like hair and clothes in morphed images with visible artifacts. Then, we blend the morphed face with an auxiliary image (background image) to remove artifacts outside the face region.

\begin{figure*}[t]
	%	\vspace{-4mm}
	\begin{center}
%	\setlength{\abovecaptionskip}{-0.01cm} %调整图片与标题距离 HGFM-v6.pdf
	\includegraphics[width=0.95\linewidth]{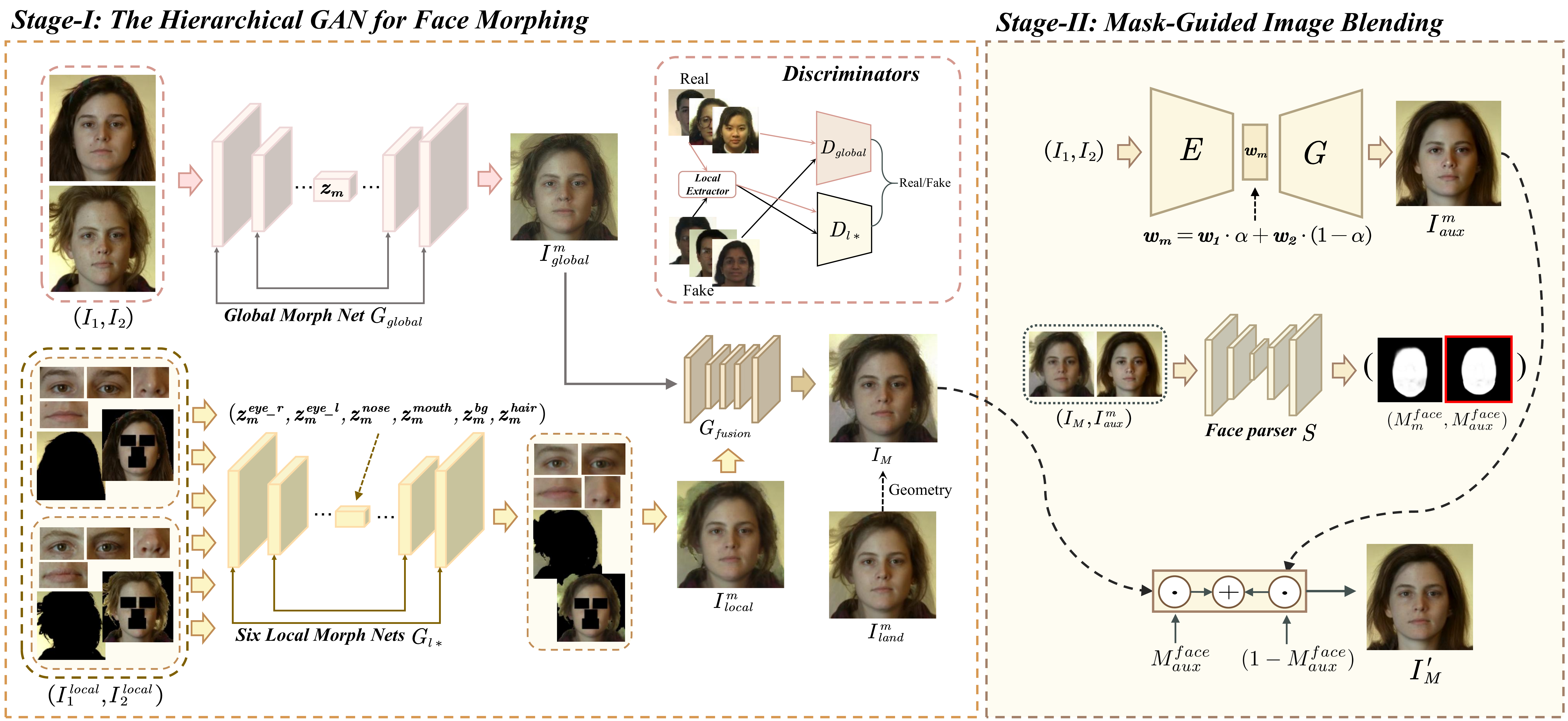} %[width=0.8\linewidth]
	\end{center}
\caption{Overview of the proposed HGFM method. It consists of two modules, the hierarchical GAN for face morphing and mask-guided image blending. The former uses hierarchical generative network containing a global morph net $G_{global}$ and six local morph nets $G_{l*}$ to generate a global morphed image $I_{global}^m$ and a local morphed image $I_{local}^m$, respectively, and then fuse them to obtain an intermediate morphed image $I_M$. The latter uses a pre-trained StyleGAN2 model to generate an auxiliary morphed image $I_{aux}^m$, and the face parser is used to define the face masks $M_{m}^{face}$ and $M_{aux}^{face}$ of $I_M$ and $I_{aux}^m$ for image blending. The final result $I_M'$ is generated by combining $I_M$ and $I_{aux}^m$ according to $M_{aux}^{face}$.}
\label{fig:short}
%\vspace{-5mm}
\end{figure*}

\section{Methodology}
\subsection{Overview of HGFM}
A high-level overview of HGFM is presented in Fig.~\ref{fig:short}. Our model uses multiple constraints to generate morphed images end-to-end through a hierarchical structure. The morphing process in HGFM can be roughly divided into the following steps:

\textbf{Initialization.} Given two bona fide images denoted as $I_1$ and $I_2$, a landmark morphed image $I_{land}^m$ is generated by OpenCV~\cite{Alpher08} morphing algorithm. This provides us with facial geometry information for morphing process. In addition, $I_1$  and ${I}_2$ are embedded into the latent space of a pre-trained StyleGAN2 encoder $E$ to obtain the corresponding latent codes $\boldsymbol{w_1}$ and $\boldsymbol{w_2}$. Then the latent codes are combined using a morph factor of 0.5 and fed into the StyleGAN2 generator $G$ to generate an auxiliary morphed image $I_{aux}^m$.

\textbf{Stage-I: Face Morphing.} At this step, the global morphed image $I_{global}^m$ and the local morphed image $I_{local}^m$ are generated by global and local morph networks, respectively. Then the fusion network $G_{fusion}$ fuses $I_{global}^m$ and $I_{local}^m$ together to obtain the intermediate morphed image $I_M$.%These networks use loss functions for backpropagation and image morphing.

\textbf{Stage-II: Mask-Guided Image Blending.} We use the face parser $S$ to get face masks $M_{m}^{face}$ and $M_{aux}^{face}$ for $I_M$ and $I_{aux}^m$. During the blending step, the $M_{aux}^{face}$ is used as the blending mask to combine $I_M$ with $I_{aux}^m$ together to generate the final morphed image $I_M'$.
\subsection{Framework Architecture}
HGFM is established on two separate modules to achieve the generation of morphed images: $(i)$ the hierarchical GAN for face morphing, and $(ii)$ a mask-guided image blending module used to eliminate visible artifacts.

\subsubsection{The Hierarchical GAN for Face Morphing}
%\label{section:3.2.1}
%\noindent \textbf{The Hierarchical GAN for Face Morphing.}
%\newline
Different from the standard GAN architecture, our proposed hierarchical structure targets both \textit{hierarchical generator} and \textit{hierarchical discriminator}, each containing a global network and six local networks. The global network analyzes and processes the holistic face image to extract global consistency information. Since the facial region contains richer identity information, we utilize six local networks to extract the subtle features of the eyes, nose, mouth, and other effective facial regions. During the morphing process, we not only perform a global fusion of the contributing identities, but also consider the varying levels of importance of different facial regions in recognition. Accordingly, we fuse contributing subjects' facial regions separately. The hierarchical structure aims to overcome the limitations of a single network by efficiently extracting features from face images with diverse regions.

\textbf{Hierarchical Generator.} In the hierarchy of $G=\{G_{global}, G_{l*}, G_{fusion}\}$, $G_{global}$ is the global generator, $G_{l*}=\{G_{eye\_l}, G_{eye\_r}, G_{nose}, G_{mouth}, G_{bg}, G_{hair}\}$ is a set of six local generators, and  $G_{fusion}$ is a fusion network. The U-Net structure is used to build all of the generators. $G_{global}$ is a U-Net with eight down-convolutional and eight up-convolutional blocks. It extracts similar shape and texture features from the contributing images to fuse them effectively. Each of  $G_{eye\_l}$, $G_{eye\_r}$, $G_{nose}$, and $G_{mouth}$ is a U-Net with three down-convolutional and three up-convolutional blocks. We use MTCNN~\cite{Alpher13} to detect the central landmarks of each facial region (\textit{i.e.}, lefit eye, right eye, nose and mouth). Based on these landmarks, we extract pairs of facial region patches from the image pair (${I}_1$, ${I}_2$). For each patch pair, we use a linear combination of them to obtain the average facial region patch $(\boldsymbol{z}_m^{eye\_l}, \boldsymbol{z}_m^{eye\_r}, \boldsymbol{z}_m^{nose}, \boldsymbol{z}_m^{mouth})$. Then, we feed them into the corresponding local generator $G_{l*}$ to generate a set of intermediate states for local morphed patches. Each local generator captures detailed information on facial regions such as eye color, mouth shape and so on. This compensates for the limitations of the global network, which cannot capture such fine-grained details. 
Each of $G_{bg}$ and $G_{hair}$ is a U-Net with four down-convolutional and four up-convolutional blocks. The background regions in ${I}_1$ and ${I}_2$ are detected using the portrait segmentation method~\cite{Alpher14} and interpolated to obtain $\boldsymbol{z}_m^{bg}$, which is input to $G_{bg}$. The remaining region of ${I}_1$ and ${I}_2$ are interpolated together to obtain $\boldsymbol{z}_m^{hair}$ as an input to $G_{hair}$. All the outputs of local generators are combined to make a complete local morphed image $I_{local}^m$. This image reattaches facial region patches based on the central landmarks detected by MTCNN~\cite{Alpher13} and addresses overlapping regions through minimum pooling. Additionally, the generator includes a fusion network $G_{fusion}$, which comprises a flat convolutional module block, three residual blocks, and a final convolutional layer. This network is responsible for synthesizing the global and local morphed images, $I_{global}^m$ and $I_{local}^m$, to generate the intermediate morphed image $I_M$.

\textbf{Hierarchical Discriminator.} The role of the discriminator is to distinguish between real and fake input images. In the hierarchy of $D=\{D_{global}, D_{l*}\}$, $D_{global}$ is a global discriminator, and $D_{l*}=\{D_{eye\_l}, D_{eye\_r}, D_{nose}, D_{mouth}, D_{bg}, D_{hair}\}$ is a set of six local discriminators. $D_{global}$ uses multi-scale discriminator in Pix2PixHD~\cite{Alpher16} to examine the entire image and extract multiple layers of features from the real and fake images. $D_{l*}$ uses Markovian discriminator in Pix2Pix~\cite{Alpher15} to examine distinct local facial regions to evaluate the quality of fine details. The Markovian discriminator process every $70\times70$ patch of the input image and evaluates the style of each patch, enabling the discriminator to learn local patterns at various levels of granularity, such as coarse and fine levels of the local input. 
This procedure can better distinguish bona fide images from morphed images.

\subsubsection{The Mask-Guided Image Blending}
%\noindent \textbf{The Mask-Guided Image Blending.}
%Since the morphed image generated by the first module has noticeable artifacts in hair, ear, neck, and other areas, which can be easily detected by the human observers, we use a mask-guided blending network to effectively remove the above artifacts for visibly realistic morphed images. 
Existing morphing methods, either landmark-based or generation-based methods, do not carefully consider the impact of non-face regions in contributing face images, resulting in ghost artifacts in morphed images. To address the noticeable artifacts present in areas such as hair, ears, and neck in the morphed images generated by the first module, we incorporate a mask-guided blending network. This network effectively removes these artifacts and ensures the production of visibly realistic morphed images. First, we use a pre-trained StyleGAN2 model to generate an auxiliary morphed image $I_{aux}^m$. Then, as shown in Fig.~\ref{fig:short}, we use a face parser $S$ based on DeepLabV3~\cite{chen2017rethinking} to predict the face components of the image. Formally, the model implements a mapping from an image to a tensor of probabilities along the channel dimension, \textit{i.e.}: $S: \mathbb{R}^{3 \times n \times m} \to [0,1]^{L \times n \times m}$, where $L$ is the number of face components. The face mask $M^{face}$ of image $I$ includes several effective facial regions and is calculated as follows:
%Eq.~\ref{eq:face}.
%The face mask $M_{face}$ includes facial regions such as skin, eyes, eyebrows, nose, and mouth, which are calculated as follows:
\begin{equation} \label{eq:face} \begin{aligned}
		M^{face}(I) &= S_{skin}(I) + S_{eyes}(I) + S_{nose}(I) \\
		&+ S_{eyebrows}(I) + S_{mouth}(I).
	\end{aligned}
\end{equation}

The face masks ($M_{m}^{face}$ and $M_{aux}^{face}$) are calculated individually for the intermediate morphed image $I_M$ and the auxiliary morphed image $I_{aux}^m$ by (\ref{eq:face}). We take $I_M$ as the foreground image, $I_{aux}^m$ as the background image, and $M_{aux}^{face}$ as the blending mask. The final morphed image $I_M'$ is generated as (\ref{eq: blending}). During the blending stage, the facial regions of $I_M$ and $I_{aux}^m$ may have different sizes, causing artifacts in the facial contours of $I_M'$. Therefore, we propose a mask loss (see (\ref{eq:mask})) to solve this problem.

\begin{equation}  \label{eq: blending}
	I_M' = M_{aux}^{face} \odot I_M + (1-M_{aux}^{face}) \odot I_{aux}^m.
\end{equation}

\begin{table*}[t]
	\caption{\footnotesize Comparison of MMPMR@FMR=0.1\%, FID, SSIM and PSNR values of our approach with other methods on FERET, FRLL, and FRGC datasets. FN, AF, CF, and EF denote FaceNet, ArcFace, CurricularFace, and ElasticFace, respectively. Best results are shown in bold.}
	\vspace{-11 pt}
	\label{mmpmr}
	\begin{center}
		\begin{tabular}{| c | c | c | c | c | c | c | c | c |}
			\hline \multirow{2}{*}{\textbf{Dataset}} & \multirow{2}{*}{\textbf{Method}} & \multirow{2}{*}{\textbf{FID$\downarrow$}} & \multirow{2}{*}{\textbf{SSIM$\uparrow$}} & \multirow{2}{*}{\textbf{PSNR$\uparrow$}} & \multicolumn{4}{c|}{ \textbf{MMPMR (\%)$\uparrow$}} \\
			\cline {6 - 9} & & & & & \textbf{FN}~\cite{facenet}& \textbf{AF}~\cite{arcface}&  \textbf{CF}~\cite{curricularface} & \textbf{EF}~\cite{elasticface} \\
			\hline \multirow{7}{*}{ FERET } 
			& FaceMorpher~\cite{Alpher09} & 56.5 & 0.5860 & 12.91 & 58.6 & 83.2 & 91.7 &  84.7  \\
			& OpenCV~\cite{Alpher08}&  53.8 & 0.6278 & 14.63 & 58.4 & 85.3 & \textbf{93.7} & 86.2 \\
			& StyleGAN2~\cite{Alpher03}& \textbf{38.3} & 0.6360 & 14.71 & 18.0 & 24.1 & 14.4 & 7.9  \\
			& MIPGAN-II~\cite{Alpher05} & 41.2 & 0.6411 & 14.94 & 47.3 & 53.4 & 68.4 & 54.7 \\
			& ReGenMorph~\cite{2021regenmorph} & 58.6 & \textbf{0.6921} & \textbf{16.60} & 45.7 & 77.3 & 81.7 & 72.8 \\
			& MorDIFF~\cite{2023mordiff} & 56.8 & 0.6668 & 15.93 & 47.2 & 78.6 & 83.9 &  79.9  \\
			\rowcolor{gray!30}& HGFM (Ours) & 40.5 & 0.6756 & 15.15 & \textbf{59.9} & \textbf{85.8} & 93.4 & \textbf{88.9} \\
%			&\rowcolor{gray!30} HGFM (Ours) & 40.5 & 0.6756 & 15.15 & \textbf{59.9} & \textbf{85.8} & 93.4 & \textbf{88.9} \\
			\hline \multirow{7}{*}{ FRLL } 
			& FaceMorpher~\cite{Alpher09} & 81.2 & 0.6198 & 15.10 & 92.8  &  98.4  & 99.6  & 99.4 \\
			& OpenCV~\cite{Alpher08}& 69.7 & 0.6754 & 17.01 & 90.8 & \textbf{99.5} & 99.4 &  99.6\\
			& StyleGAN2~\cite{Alpher03}& 36.1 & 0.6786 & 17.09 & 37.6 & 78.7 & 63.5 & 57.2 \\
			& MIPGAN-II~\cite{Alpher05} & 37.5 & 0.7036 & 17.41 & 82.9 & 81.9 & 90.1 & 89.9\\
			& ReGenMorph~\cite{2021regenmorph} & 58.0 & 0.7213 & \textbf{19.33} & 84.5 & 97.4 & 99.1 & 98.0\\
			& MorDIFF~\cite{2023mordiff} & 49.4 & \textbf{0.7356} & 18.73 & 82.6  &  96.7  & 98.8  & 98.8 \\
			\rowcolor{gray!30}&HGFM (Ours) & \textbf{33.7} & 0.7013 & 17.90 & \textbf{93.5} & 99.1 & \textbf{99.9} & \textbf{99.9} \\
%			&\rowcolor{gray!30} HGFM (Ours) & \textbf{33.7} & 0.7013 & 17.90 & \textbf{93.5} & 99.1 & \textbf{99.9} & \textbf{99.9} \\
			\hline \multirow{7}{*}{ FRGC } 
			& FaceMorpher~\cite{Alpher09} & 76.9 & 0.6903 & 20.30 &  82.7  &  \textbf{96.7}  & 97.8 & 96.7  \\
			& OpenCV~\cite{Alpher08}& \textbf{68.3} & 0.7208 & 20.96 & 81.0 & 96.0 & 97.6  &  96.1\\
			& StyleGAN2~\cite{Alpher03}& 73.4 & 0.6749 & 19.27 & 23.6 & 53.6 & 35.2 & 26.0 \\
			& MIPGAN-II~\cite{Alpher05} & 87.1 & 0.6739 & 19.21 & 77.0 & 77.7 & 88.2 & 81.4\\
			& ReGenMorph~\cite{2021regenmorph} & 98.5 & 0.7264 & \textbf{21.14} & 68.3 & 92.6 & 88.9 & 88.9\\
			& MorDIFF~\cite{2023mordiff} & 93.8 & 0.7339 & 20.88 & 66.2  &  95.0  & 95.8  & 94.2 \\
			\rowcolor{gray!30}& HGFM & 88.4 & \textbf{0.7408} & 20.32 & \textbf{86.2} & 96.1 & \textbf{98.9} & \textbf{98.3}\\
%			&\rowcolor{gray!30} HGFM & 88.4 & \textbf{0.7408} & 20.32 & \textbf{86.2} & 96.1 & \textbf{98.9} & \textbf{98.3}\\
			\hline
		\end{tabular}%\vspace{-10mm}
	\end{center}
\end{table*}

\subsection{Loss Function}
%During the training stage, multiple loss functions are employed to optimize the morphing process. These include $(i)$ an \textit{appearance constraint} to enhance the texture detail of the contributing subjects; $(ii)$ a \textit{geometry constraint} to ensure that the morphed images also contain the geometric facial structure of the contributors; $(iii)$ an \textit{identity--preservation constraint} to improve the robustness of the morphed images; $(iv)$ a \textit{mask constraint} to specify the size of the face region when blending is performed.
During the training stage, multiple loss terms are employed to optimize the morphing process. 
%we use multiple loss terms to optimize the morphing process, which 
These loss terms include the geometry loss $\mathcal{L}_{gm}$, the combined identity loss $\mathcal{L}_{cid}$, the mask loss $\mathcal{L}_{mask}$, the appearance loss $\mathcal{L}_{app}$, and the adversarial loss $\mathcal{L}_{adv}$.

%\noindent
\textbf{Geometry Loss.} Previous studies~\cite{2021regenmorph,Alpher03} suggest that morphed images produced by generation-based methods often lose significant facial geometric information, whereas landmark-based methods preserve the geometric information of the contributing subjects well. Being aware of this, we propose to use landmark-based morphed images to supervise the generation of morphed images at the pixel-level. By leveraging the rich information contained in the landmark-based morphed images to guide the creation of morphed images with greater fidelity. The geometry loss can be written as below.
\begin{equation} \label{eq:geometry} 
	\mathcal{L}_{gm} = \Vert I_M - I_{land}^m \Vert_1,
\end{equation}
where we use the $L_1$ norm to measure the pixel loss between the intermediate morphed image and the landmark morphed image.

%\noindent
\textbf{Combined Identity Loss.} 
%To make morphed images that can effectively attack FRSs, However, it may not be practical for unknown FRSs.
Previous studies~\cite{Alpher02,Alpher05} only used a single pre-trained face recognition (FR) model to calculate the identity loss. Unlike the standard identity loss, we propose to use multiple FR models to jointly calculate the identity loss between the intermediate morphed image $I_M$ and the two contributing images, using ArcFace~\cite{arcface} and FaceNet~\cite{facenet}, respectively. This strategy aims to improve the generalization ability of our morphing method under known and unknown FR techniques. 
%To improve the generalization ability of our model to different FR technologies, we propose to utilize multiple FR models to calculate the identity loss between the intermediate morphed image $I_M$ and the two contributing images, using ArcFace~\cite{arcface} and FaceNet~\cite{facenet}, respectively.
The combined identity loss is formulated as follows:
\begin{equation}\label{eq:identity}
	\begin{aligned}
		\mathcal{L}_{cid} = \frac{d(\boldsymbol{z}_m^a, \boldsymbol{z}_1^a)+d(\boldsymbol{z}_m^a, \boldsymbol{z}_2^a)}{2} + \frac{d(\boldsymbol{z}_m^f, \boldsymbol{z}_1^f)+d(\boldsymbol{z}_m^f, \boldsymbol{z}_2^f)}{2},
	\end{aligned}
\end{equation}
where $d(\cdot)$ denotes the cosine distance function, $\boldsymbol{z}_*^a$ and $\boldsymbol{z}_*^f$ are the features extracted by ArcFace~\cite{arcface} and FaceNet~\cite{facenet} respectively. 

\textbf{Mask Loss.} Differences in facial region sizes between the intermediate morphed image $I_M$ and the auxiliary morphed image $I_{aux}^m$ can result in visible artifacts around the face during blending, impacting image quality. To address this, we introduce the mask loss following~\cite{2021high}.
%Possible facial region size inconsistencies between the intermediate morphed image $I_M$ and the auxiliary morphed image $I_{aux}^m$ lead to visible artifacts around the face during the blending step and affect the perceived quality of the image. To avoid this, we introduce the mask loss $\mathcal{L}_{mask}$, which ensures that the face region size in $I_M$ is consistent with $I_{aux}^m$. The mask loss is formulated as below.
%where $M^{face}(*)$ denotes the face mask using Eq.~\ref{eq:face}.
%The face parser $S$ is used to define this constraint, and the mask loss is formulated as follows:
\begin{equation} \label{eq:mask}
	\mathcal{L}_{mask} = \Vert M_{m}^{face}-M_{aux}^{face}\Vert_2^2.
\end{equation}

\textbf{Appearance Loss.} To enrich the texture details of the contributing images within the morphed image, we employ an appearance loss comprising a perceptual loss~\cite{Alpher05}, a weak feature matching loss~\cite{Alpher21}, and a local loss~\cite{yi2019apdrawinggan}. The loss function is formulated as below.

%\begin{equation}
%	\begin{aligned}
	%		\mathcal{L}_{per}= \frac{1}{2}\left\|H(I_1)-H(I_M)\right\|_2^2 +\left\|H(I_2)-H(I_M)\right\|_2^2 ,\nonumber
	%	\end{aligned}
%\end{equation}
%
%\begin{equation}
%	\begin{aligned}
	%		\mathcal{L}_{wfm}= \sum_{i=1}^2 \frac{1}{N_i}\left\|D^{(i)}\left(I_M\right)-D^{(i)}\left(I_{land}^m\right)\right\|_1,\nonumber
	%	\end{aligned}
%\end{equation}
%\begin{equation}
%	\begin{aligned}
	%		\mathcal{L}_{local} = \sum_{i=1}^6 \Vert P_i(I_M) - P_i(I_{land}^m) \Vert_1,\nonumber
	%	\end{aligned}
%\end{equation}
\begin{equation}\label{eq:app}
	\begin{aligned}
		\mathcal{L}_{app} = \mathcal{L}_{per} + \mathcal{L}_{wfm} + \mathcal{L}_{local}.
	\end{aligned}
\end{equation}
%where $H(\cdot)$ denotes the perceptual feature extractor, $D$ is the multi-scale discriminator ($D_{global}$), and $P_i(I)$ denotes the \emph{i-th} face region of image $I$ (six regions in total).

%\noindent
\textbf{Adversarial Loss.} 
%The hierarchical discriminator contains one global discriminator and six local discriminators. 
The adversarial loss term consists of two parts: the Hinge version adversarial loss for global adversarial loss $\mathcal{L}_{adv}^{global}$ and the BCE version adversarial loss for local adversarial loss $\mathcal{L}_{adv}^{local}$. The adversarial loss is formulated as below.
%\mathop{\min}_{G_{global}} \mathop{\max}_{D_{global}}
\begin{equation}\label{global_adv}
	\begin{aligned}
		\mathcal{L}_{adv}^{global} &=\mathbb{E}_{\boldsymbol{x} \sim X}[\max (0,1-D(\boldsymbol{x}))] \\ &+\mathbb{E}_{\boldsymbol{y} \sim Y}[\max (0,1+D(\boldsymbol{y}))]-\mathbb{E}_{\boldsymbol{y} \sim Y}[D(\boldsymbol{y})] ,\nonumber
	\end{aligned}
\end{equation}

\begin{equation}\label{local_adv}
	\begin{aligned}
		\mathcal{L}_{adv}^{local} &= \sum_{D_j \in D_{l*}} \mathbb{E}_{(\boldsymbol{x}_i) \sim X}[\operatorname {log}(D_j(\boldsymbol{x}_i))] \\
		&+ \mathbb{E}_{(\boldsymbol{y}_i) \sim Y}[\log(1-D_j(\boldsymbol{y}_i))],\nonumber
	\end{aligned}
\end{equation}

\begin{equation} \label{adv}
	\mathcal{L}_{adv} = \mathcal{L}_{adv}^{global} + \mathcal{L}_{adv}^{local},
\end{equation}
where $\boldsymbol{x}$ is the real images, $\boldsymbol{y}$ is the generated morphed images, $\boldsymbol{x}_i$ is the \textit{i-th} face region of the real image, and $\boldsymbol{y}_i$ is the \textit{i-th} face region of the morphed image.
%where $\boldsymbol{x}$ is the bona fide images, and $\boldsymbol{y}$ is the generated morphed images.

Thus, the overall loss function can be formulated as:
\begin{equation}\label{sum} 
	\begin{split}
		\mathcal{L}_{all} &= \lambda_1 \mathcal{L}_{gm} + \lambda_2 \mathcal{L}_{cid} + \lambda_3 \mathcal{L}_{mask} + \lambda_4 \mathcal{L}_{app} + \lambda_5 \mathcal{L}_{adv},
	\end{split}
\end{equation}
where $\lambda_1$, $\lambda_2$, $\lambda_3$, $\lambda_4$ and $\lambda_5$ are the hyper-parameters that are set to achieve stable and generalized convergence. In this work, we set  $\lambda_1, \lambda_2, \lambda_3, \lambda_4 = 10$, $\lambda_5 = 1$, respectively.

\begin{table*}[t]
	%	\vspace{-6mm}
	\caption{Detection results of MAD methods MixFaceNet (MFN) and HRNet (HRN) on FERET dataset. Both models are trained on the SMDD dataset. Higher is better.}
	%\vspace{-20pt}
	\vspace{-11 pt}
	\label{mad}
	%	\smallskip
	\begin{center}
		%		\resizebox{\columnwidth}{!}{%
			\begin{tabular}{  |c | c | c | c | c | c | c | c | c | c | c |}
				\hline  
				\multirow{2}{*}{\textbf{MAD}} & \multirow{2}{*}{\textbf{Method}} & \multirow{2}{*}{\textbf{EER (\%)}} &
				\multicolumn{4}{c|}{\textbf{APCER(\%)@BPCER =}} & 
				\multicolumn{4}{c|}{\textbf{BPCER(\%)@APCER =}} \\
				\cline{4-11}
				& & &1\% & 5\%  & 10\% & 20\% & 1\% & 5\% & 10\% & 20\%  \\ \hline
				\multirow{7}{*}{MFN~\cite{smdd}} 
				& FaceMorpher~\cite{Alpher09} & 9.64 & 44.23 & 16.07 & 8.32 & 2.27 & 24.15 & 16.23 & 8.68 & 4.15 \\
				& OpenCV~\cite{Alpher08} & 13.23 & 65.97 & 36.67 & 23.06 & 6.81 & 38.87 & 21.32 & 15.85 & 11.13 \\
				& StyleGAN2~\cite{Alpher03} & 24.76 & 82.80 & 65.41 & 50.42 & 30.24 & 77.36 & 55.28 & 41.89 & 29.62\\
				& MIPGAN-II~\cite{Alpher05} & 30.81 & 79.21 & 63.52 & 53.31 & 39.51 & 83.39 & 65.09 & 45.66 & 35.28\\
				& ReGenMorph~\cite{2021regenmorph} & 11.91 & 43.86 & 20.79 &10.59 & 1.70 & 24.72 & 15.47 & 11.13 & 5.47\\
				& MorDIFF~\cite{2023mordiff} & 37.81 & 92.44 & 81.85 & 70.89 & 55.20 & 86.60 & 75.28 & 66.23 & 50.19\\
				\rowcolor{gray!30}&HGFM (Ours) & \textbf{39.31} & \textbf{96.03} & \textbf{88.85} & \textbf{79.21} & \textbf{63.89} & \textbf{90.19} & \textbf{74.91} & \textbf{66.04}& \textbf{54.34}\\\hline
%			    & \rowcolor{gray!30} HGFM (Ours) & \textbf{39.31} & \textbf{96.03} & \textbf{88.85} & \textbf{79.21} & \textbf{63.89} & \textbf{90.19} & \textbf{74.91} & \textbf{66.04}& \textbf{54.34}\\\hline
				%				\multirow{5}{*}{XN~\cite{2022syn}} 
				%				& FaceMorpher & 11.37 & 2.62 & 1.39 & 0.33 & 10.78 & 1.47 & 0.98 & 0.49 \\
				%				& OpenCV & 16.54 & 2.95 & 1.64 & 0.33 & 11.27 & 1.96 & 1.47 & 0.49 \\
				%				& StyleGAN2 & 73.98 & 39.85 & 25.94 & 14.73 & 100.0 & 43.63 & 29.41 & 11.76\\
				%				& MIPGAN-II  & 58.55 & 26.65 & 18.13 & 9.84 &100.0 & 30.88 & 19.61 & 6.86\\
				%				& HGFM & 78.46 & 25.55 & 14.17 & 5.16 & 35.29 & 20.59 & 11.27 & 5.39\\\hline 
				\multirow{7}{*}{HRN~\cite{2022syn}} 
				& FaceMorpher~\cite{Alpher09} & 10.02 & 70.32 & 28.17 & 9.83 & 0.76 & 19.43 & 13.77 & 10.00 & 6.23 \\
				& OpenCV~\cite{Alpher08} & 14.18 & 86.77 & 48.96 & 27.98 & 4.72 & 32.26 & 20.00 & 16.04 & 12.07 \\
				& StyleGAN2~\cite{Alpher03} & 28.54 & 93.38 & 78.07 & 68.81 & 44.42 & 100.0 & 100.0 & 44.15 & 33.58 \\
				& MIPGAN-II~\cite{Alpher05}  & 23.06 & 89.79 & 69.56 & 54.06 & 32.14 & 100.0 & 100.0 & 38.11 & 26.41\\
				& ReGenMorph~\cite{2021regenmorph}  & 8.88 & 73.72 & 25.52 & 6.81 & 0.76 & 19.62 & 12.64 & 8.68 & 6.04\\
				& MorDIFF~\cite{2023mordiff} & 34.78 & 96.03 & 86.77 & 77.32 & 59.92 & 100.0 & 100.0 & 100.0 & 43.40\\
				\rowcolor{gray!30}&HGFM (Ours) & \textbf{35.54} & \textbf{99.43} & \textbf{90.37} & \textbf{82.41} & \textbf{65.03} & \textbf{100.0} & \textbf{100.0} & \textbf{100.0} & \textbf{46.04} \\\hline   
%				&\rowcolor{gray!30} HGFM (Ours) & \textbf{35.54} & \textbf{99.43} & \textbf{90.37} & \textbf{82.41} & \textbf{65.03} & \textbf{100.0} & \textbf{100.0} & \textbf{100.0} & \textbf{46.04} \\\hline   
			\end{tabular}%}\vspace{-4mm}
	\end{center}
\end{table*}

\begin{table*}[ht]
	%	\vspace{-6mm}
	\caption{Detection results of MAD methods MixFaceNet (MFN) and HRNet (HRN) on FRLL dataset.}
	%	\vspace{-20pt}
	\vspace{-11 pt}
	\label{mad2}
	%	\smallskip
	\begin{center}
		%		\resizebox{\columnwidth}{!}{%
			\begin{tabular}{  |c | c | c | c | c | c | c | c | c | c | c |}
				\hline  
				\multirow{2}{*}{\textbf{MAD}} & \multirow{2}{*}{\textbf{Method}} & 
				\multirow{2}{*}{\textbf{EER (\%)}} &
				\multicolumn{4}{c|}{\textbf{APCER(\%)@BPCER =}} & 
				\multicolumn{4}{c|}{\textbf{BPCER(\%)@APCER =}} \\
				\cline{4-11}
				& & & 1\% & 5\%  & 10\% & 20\% & 1\% & 5\% & 10\% & 20\%  \\ \hline
				\multirow{7}{*}{MFN~\cite{smdd}} 
				& FaceMorpher~\cite{Alpher09} & 2.29 & 2.29 & 1.72 & 1.23 & 0.65 & 13.24 & 1.47 & 1.47 & 1.47\\
				& OpenCV~\cite{Alpher08} & 1.97 & 2.29 & 1.31 & 0.98 & 0.66 & 12.75 & 0.49 & 0.49 & 0.49 \\
				& StyleGAN2~\cite{Alpher03} & \textbf{12.85} & 56.55 & 29.54 & 15.96 & 6.96 & 50.98 & 26.96 & 15.69 & 6.86\\
				& MIPGAN-II~\cite{Alpher05} & 2.21 & 9.17 & 1.80 & 1.39 & 1.23 & 30.88 & 2.45& 0.49 & 0.49\\
				& ReGenMorph~\cite{2021regenmorph} & 1.15 & 1.06 & 0.25 & 0.16 & 0.16 & 1.47 & 1.47 & 1.47 & 1.47\\
				& MorDIFF~\cite{2023mordiff} & 6.97 & 45.04 & 11.24 & 4.68 & 3.12 & 59.80 & 8.33 & 4.90 & 2.94\\
				\rowcolor{gray!30}&HGFM (Ours) & 12.78 & \textbf{76.58}  & \textbf{34.56} & \textbf{16.72} & \textbf{7.53} & \textbf{77.94} & \textbf{32.35} & \textbf{16.18} & \textbf{7.35}\\\hline
%				&\rowcolor{gray!30} HGFM (Ours) & 12.78 & \textbf{76.58}  & \textbf{34.56} & \textbf{16.72} & \textbf{7.53} & \textbf{77.94} & \textbf{32.35} & \textbf{16.18} & \textbf{7.35}\\\hline
				%				\multirow{5}{*}{XN~\cite{2022syn}} 
				%				& FaceMorpher & 11.37 & 2.62 & 1.39 & 0.33 & 10.78 & 1.47 & 0.98 & 0.49 \\
				%				& OpenCV & 16.54 & 2.95 & 1.64 & 0.33 & 11.27 & 1.96 & 1.47 & 0.49 \\
				%				& StyleGAN2 & 73.98 & 39.85 & 25.94 & 14.73 & 100.0 & 43.63 & 29.41 & 11.76\\
				%				& MIPGAN-II  & 58.55 & 26.65 & 18.13 & 9.84 &100.0 & 30.88 & 19.61 & 6.86\\
				%				& HGFM & 78.46 & 25.55 & 14.17 & 5.16 & 35.29 & 20.59 & 11.27 & 5.39\\\hline 
				\multirow{7}{*}{HRN~\cite{2022syn}} 
				& FaceMorpher~\cite{Alpher09} & 0.98 & 0.90 & 0.49 & 0.33 & 0.24 & 0.98 & 0.49 & 0.49 & 0.49 \\
				& OpenCV~\cite{Alpher08} & 1.31 & 1.31 & 0.82 & 0.49 & 0.24 & 4.41 & 0.49 & 0.49 & 0.49 \\
				& StyleGAN2~\cite{Alpher03} & \textbf{13.67} & \textbf{38.62} & 25.23 & \textbf{20.46} & 7.77 & 42.65 & 24.51 & \textbf{16.67} & 8.80 \\
				& MIPGAN-II~\cite{Alpher05}  & 9.57 & 23.57 & 16.84 & 11.14 & 3.37 & 45.10 & 15.69 & 10.29 & 2.94\\
				& ReGenMorph~\cite{2021regenmorph}  & 8.94 & 3.19 & 0.24 & 0.24 & 0.08 & 2.45 & 2.45 & 2.45 & 0.00\\
				& MorDIFF~\cite{2023mordiff}  & 0.33 & 3.19 & 1.06 & 0.66 & 0.16 & 6.37 & 0.49 & 0.49 & 0.49\\
				\rowcolor{gray!30}& HGFM (Ours)& 11.96 & 37.18 & \textbf{25.47} & 18.59 & \textbf{8.35} & \textbf{53.43} & \textbf{27.45} & 15.20 & \textbf{8.82} \\\hline  
%				&\rowcolor{gray!30} HGFM (Ours)& 11.96 & 37.18 & \textbf{25.47} & 18.59 & \textbf{8.35} & \textbf{53.43} & \textbf{27.45} & 15.20 & \textbf{8.82} \\\hline 
			\end{tabular}%}\vspace{-4mm}
	\end{center}
\end{table*}

\begin{table*}[ht]
	%	\vspace{-6mm}
	\caption{Detection results of MAD methods MixFaceNet (MFN) and HRNet (HRN) on FRGC dataset.}
	%	\vspace{-20pt}
	\vspace{-11 pt}
	\label{mad3}
	%	\smallskip
	\begin{center}
		%		\resizebox{\columnwidth}{!}{%
			\begin{tabular}{  |c | c | c | c | c | c | c | c | c | c | c |}
				\hline  
				\multirow{2}{*}{\textbf{MAD}} & \multirow{2}{*}{\textbf{Method}} & 
				\multirow{2}{*}{\textbf{EER (\%)}} &
				\multicolumn{4}{c|}{\textbf{APCER(\%)@BPCER =}} & 
				\multicolumn{4}{c|}{\textbf{BPCER(\%)@APCER =}} \\
				\cline{4-11}
				& & & 1\% & 5\%  & 10\% & 20\% & 1\% & 5\% & 10\% & 20\%  \\ \hline
				\multirow{7}{*}{MFN~\cite{smdd}} 
				& FaceMorpher~\cite{Alpher09} & 75.41 &100.0  & \textbf{99.89} & \textbf{99.69} &97.72 & 100.0 & 98.97 &98.77  &97.13 \\
				& OpenCV~\cite{Alpher08} & 71.72 &100.0  & 99.79 & 99.48 & 95.95 & 100.0 & 98.97 & 97.54 &95.69 \\
				& StyleGAN2~\cite{Alpher03} & \textbf{83.19} & 100.0 & 99.79 & 99.27 & \textbf{98.23} &100.0  & 100.0 & \textbf{100.0} &\textbf{99.59} \\
				& MIPGAN-II~\cite{Alpher05} & 65.43 & 99.78 & 96.72 & 94.97 & 87.09 & 100.0 & 100.0 & 100.0 & 98.91 \\
				& ReGenMorph~\cite{2021regenmorph} & 1.23& 45.67 & 16.05 & 9.87 &12.3  & 30.49 & 15.85 & 13.41 & 2.44 \\
				& MorDIFF~\cite{2023mordiff} & 3.94 & 83.15 & 64.99 & 48.58 & 24.94 & 66.59 &38.65  & 32.53 & 23.14\\
				\rowcolor{gray!30}&HGFM (Ours) & 73.14 & \textbf{100.0} & 99.56 & 98.91 & 95.40 & \textbf{100.0} & \textbf{100.0} & 99.34 & 97.60\\\hline
%				&\rowcolor{gray!30} HGFM (Ours) & 73.14 & \textbf{100.0} & 99.56 & 98.91 & 95.40 & \textbf{100.0} & \textbf{100.0} & 99.34 & 97.60\\\hline
				%				\multirow{5}{*}{XN~\cite{2022syn}} 
				%				& FaceMorpher & 11.37 & 2.62 & 1.39 & 0.33 & 10.78 & 1.47 & 0.98 & 0.49 \\
				%				& OpenCV & 16.54 & 2.95 & 1.64 & 0.33 & 11.27 & 1.96 & 1.47 & 0.49 \\
				%				& StyleGAN2 & 73.98 & 39.85 & 25.94 & 14.73 & 100.0 & 43.63 & 29.41 & 11.76\\
				%				& MIPGAN-II  & 58.55 & 26.65 & 18.13 & 9.84 &100.0 & 30.88 & 19.61 & 6.86\\
				%				& HGFM & 78.46 & 25.55 & 14.17 & 5.16 & 35.29 & 20.59 & 11.27 & 5.39\\\hline 
				\multirow{7}{*}{HRN~\cite{2022syn}} 
				& FaceMorpher~\cite{Alpher09} & 8.19 & \textbf{69.09} & \textbf{16.29} & 7.26 & \textbf{1.56} &  \textbf{100.0} & 13.11 & 7.58 & \textbf{4.30}\\
				& OpenCV~\cite{Alpher08} & 7.88 & 63.17 & 10.79 & 3.53 & 1.14 & 100.0 & 8.20 & 5.33 & 3.48\\
				& StyleGAN2~\cite{Alpher03} & 6.64 & 42.32 & 8.30 & 4.05 & 0.52 & 19.26 & 8.20 & 4.30 & 2.46\\
				& MIPGAN-II~\cite{Alpher05}  & \textbf{9.41}  & 35.01 & 12.03 & \textbf{8.75} & 2.19 & 100.0 & \textbf{13.54} & \textbf{8.30} & 2.62\\
				& ReGenMorph~\cite{2021regenmorph}  & 1.23 & 24.69 &0.00  & 0.00 & 0.00 &3.66  & 2.44 & 2.44 &1.22 \\
				& MorDIFF~\cite{2023mordiff}  & 3.94 & 20.35 & 2.19 & 0.00 & 0.00 & 7.42 & 3.71 & 1.75 &1.09 \\
				\rowcolor{gray!30}& HGFM (Ours)& 0.88 & 1.31 & 0.00 & 0.00 & 0.00 & 1.31 & 0.87 & 0.87 &0.65 \\\hline  
%				&\rowcolor{gray!30} HGFM (Ours)& 0.88 & 1.31 & 0.00 & 0.00 & 0.00 & 1.31 & 0.87 & 0.87 &0.65 \\\hline  
			\end{tabular}%}\vspace{-4mm}
	\end{center}
\end{table*}

\section{Experiments}
\subsection{Implementation datails}
For each face image, we extract 68 landmarks and then align and crop the face images according to the FFHQ~\cite{Alpher04} process, with input image size $256\times256$. Our model is trained on one GeForce RTX 2080 Ti with batch size 1. We use an Adam optimizer with a fixed learning rate of $2\times 10^{-4}$ to optimize the network.

\subsection{Datasets and Evaluation Metrics}

\textbf{Datasets.} In this paper, we use the FERET~\cite{feret}, FRLL~\cite{frll}, and FRGC~\cite{frgc} face datasets to generate morphed images. To compare with previous work, we employ FERET-Morphs, FRLL-Morphs, and FRGC-Morphs~\cite{morphs}, each morph dataset contains four types of morphing attacks, namely FaceMorpher~\cite{Alpher09}, OpenCV~\cite{Alpher08}, StyleGAN2~\cite{Alpher03} and MIPGAN-II~\cite{Alpher05}. The number of images generated by each morphing method is the same as in~\cite{smdd}. In order to be consistent among different morphing methods, we select the same morphing pairs as in~\cite{morphs} in our experiments. Each face image is pre-processed following FFHQ~\cite{Alpher04} for fair comparison. 

\textbf{Evaluation Metrics.} To evaluate the threat of each morphing attack to FRSs, the mated morph presentation match rate (MMPMR)~\cite{2017biometric} based on the decision threshold at the false match rate (FMR) of 0.1\% is employed. Specifically, the thresholds are calculated on the LFW dataset.
 \begin{equation}
	MMPMR(\tau)=\frac{1}{M} \cdot \sum_{m=1}^M\{[\min _{n=1, \ldots, N_m} S_m^n]>\tau\},
\end{equation}
where $M$ and $N_m$ denote the number of morphed images and contributing images, respectively. $S_m^n$ is the similarity score for mated morph for morph $m$ of the $n$-th subject, and $\tau$ is the threshold of the FRS at a chosen FMR. Obviously, the larger the MMPMR indicates stronger threat of the morphs.

%In addition, we compute the Attack Presentation Classification Error Rate (APCER) and the Bona fide Presentation Classification Error Rate (BPCER)~\cite{2023diffusion} to assess the detectability of morphs on different morphing attack detection (MAD) algorithms. Furthermore, when APCER is equal to BPCER, the calculated Equal Error Rate (EER) is also reported. 
The performance of different morphing attack detection (MAD) algorithms is measured by the Attack Presentation Classification Error Rate (APCER, the proportion of attack images misclassified as bona fide images) and the Bona fide Presentation Classification Error Rate (BPCER, the proportion of bona fide images misclassified as attack images)~\cite{venkatesh2021face}. As it is impossible to optimize both APCER and BPCER jointly, we report results for APCER at specific BPCER values (1\%, 5\%, 10\%, 20\%) and for BPCER at specific APCER values (1\%, 5\%, 10\%, 20\%). In addition, when APCER is equal to BPCER, the Equal Error Rate (EER) is also reported. 

To quantify the visual quality of morphed images, we follow~\cite{Alpher05} and~\cite{2023diffusion} to use the Fréchet Inception Distance (FID), Structural Similarity Index (SSIM) and Peak Signal-to-Noise Ratio (PSNR) metrics. The FID metric measures the distance between the distribution of morphing attack images and the distribution of bona fide images used for the morphing attack. We use $\mathtt{pytorch\mbox{-}fid}$~\cite{pytorchfid} to compute FID. A lower FID means a closer resemblance between the distribution of generated morphing attack images and bona fide images, indicating higher visual fidelity~\cite{lucic2018gans}.

\begin{figure}[h]
	\centering
	\subfigure[FERET examples]
	{\includegraphics[width=\linewidth]{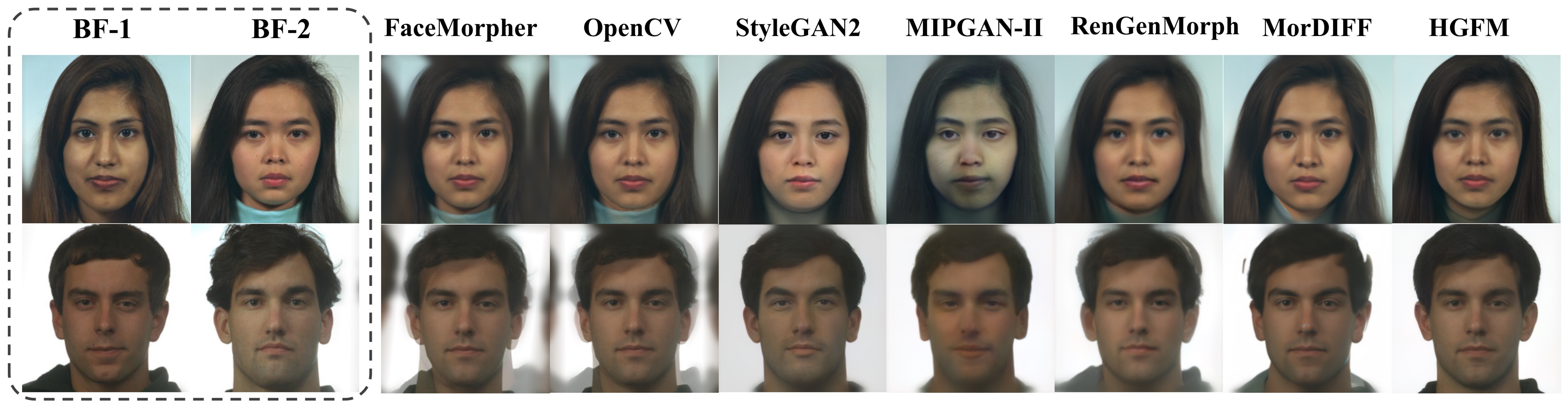}}
	
	\subfigure[FRLL examples]
	{\includegraphics[width=\linewidth]{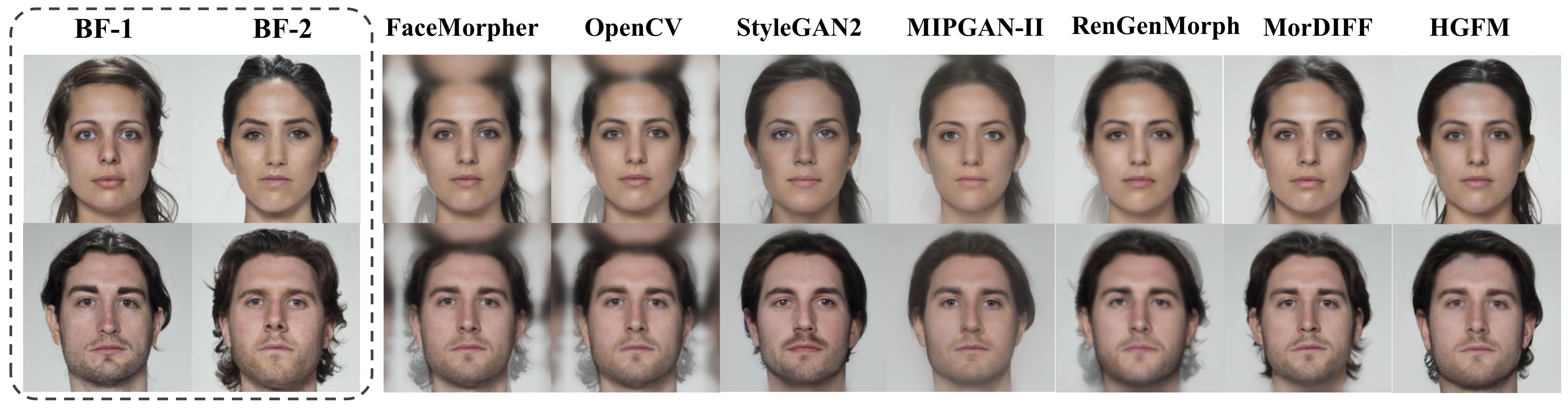}}
	
	\subfigure[FRGC examples]
	{\includegraphics[width=\linewidth]{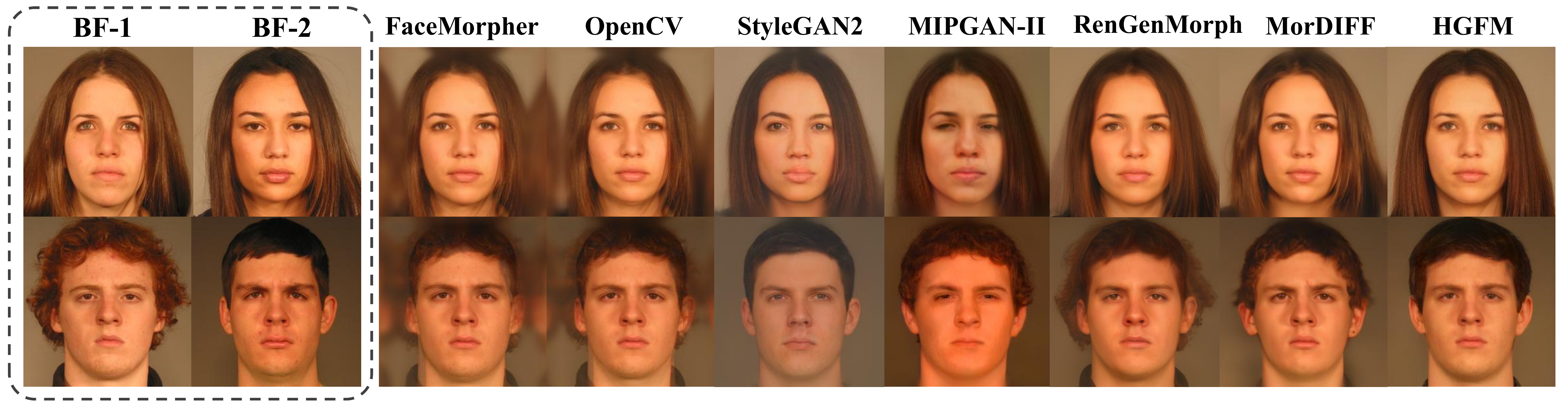}}
	
%	\caption{A qualitative result on two benchmarks. Please zoom in to see the details.}
	\caption{Examples of morphs generated by different morphing attack methods on three benchmarks. Please zoom in to see the differences among the images, especially in regions susceptible to artifacts such as the eyes, nose, mouth, and hair.}
	
	\label{fig:examples}
\end{figure}

\subsection{Analysis on Results}
\textbf{Vulnerability of FRSs.} We investigate the vulnerability of four state-of-the-art FR models, FaceNet~\cite{facenet}, ArcFace~\cite{arcface}, CurricularFace~\cite{curricularface} and ElasticFace-Arc~\cite{elasticface}, to morphing attacks. The former two FRSs use the official pre-trained models and the latter two FRSs use ResNet100 as backbone trained on MS1MV dataset. To facilitate a more effective comparison of the methods in this paper, apart from the four previously mentioned morphing methods, we have also replicated two innovative morphing methods, ReGenMorph~\cite{2021regenmorph} and MorDIFF~\cite{2023mordiff}, using the official code provided to us. 
%Besides, we reproduce ReGenMorph following the morphing procedure of~\cite{2021regenmorph}. And we obtain the MorDIFF morphed images by training the diffusion autoencoders for morphing using the official code mentioned in~\cite{2023mordiff}.
As tabulated in Table~\ref{mmpmr}, our HGFM achieves the highest MMPMR in most cases compared to other morphing methods. This is primarily attributed to the local networks capturing information that the global network overlooks, effectively bridging the information gap. Additionally, the combined identity and geometry losses effectively preserve the identity of contributing individual faces, resulting in more threatening morphed images.

%This is mainly due to the fact that the global-and-local joint morphing networks together with enhanced identity and geometry losses to better preserve the identity of contributing individual faces. That makes the morphed images more threatening.

%This is mainly due to the fact that the information captured by the local networks fills the information gap ignored by the global network, while the identity loss is calculated using two FR models, which makes the morphed images more threatening. 

\textbf{Detectability of Morphing Attacks.} To measure the detectability of our method and other baselines, we deploy two well-performing MAD algorithms, MixFaceNet~\cite{smdd} and HRNet~\cite{2022syn}. Both of them all ranked high in the SYN-MAD 2022~\cite{2022syn} competition. Lacking public code for HRNet, we train it on the SMDD dataset following the same strategy and parameters of~\cite{smdd}. For MixFaceNet, we utilize the official model for testing. Table~\ref{mad} presents the morphing attack (detectability) results on the FERET dataset. Notably the training and test sets of the MADs are disjoint, consistent with the cross-dataset evaluation in realistic scenarios. 
%Table~\ref{mad} presents the MAD performance of EER (\%), APCER(\%)@BPCER = (1\%, 10\%, 15\%, 20\%), and BPCER(\%)@APCER = (1\%, 10\%, 15\%, 20\%) on FERET dataset. 
Among all morphing methods, our HGFM ranks first against the detection methods MixFaceNet and HRNet. This indicates that our method outperforms existing state-of-the-art morphing methods in multiple attack scenarios. The detection results for the FRLL dataset are available in Table~\ref{mad2}. The results on FRGC~\cite{frgc} dataset, which is well-known for its rich image variations in illumination and expression are presented in Table~\ref{mad3}. The results show that our method performs poorly. We guess that the relatively poor MAD detection performance of our HGFM is because our current method is tailored for ICAO compliant images, but not very robust to image variations~\cite{zhang2023morphganformer,morphs,2023diffusion}.
Additionally, experiments on the quantitative detection of morphed images by human observers are available in the Supplementary Material.

%The detection results for the FRLL dataset are available in the supplementary materials. 
%\begin{figure}
%%	\vspace{-7mm}
%	\centering
%	\includegraphics[width=1\linewidth]{examples_v3.pdf}
%%	\vspace{-4mm}
%	\caption{Qualitative results on two benchmarks. Please zoom in to see the details.}
%	\label{fig:examples}
%%	\vspace{-7mm}
%\end{figure}

\begin{table*}[t]
	%	\vspace{-5mm}
	\caption{Detectability of morphed images with HRNet MAD algorithm on FERET dataset. The first row indicates that no geometry loss from (\ref{eq:geometry}) is employed, the second row indicates that only ArcFace~\cite{arcface} is used to calculate the identity loss, the third row shows that no local networks (including $G_{l*}$ and $D_{l*}$) and local loss from (\ref{eq:app}) are used, the fourth row indicates that no mask-guided image blending module and mask loss from (\ref{eq:mask}), and the fifth row indicates the replacement of mask-guided image blending with the Poisson image blending~\cite{Alpher12} method. }
	\vspace{-11pt}
	%	\vspace{-11 pt}
	\label{astudy}
	%	\smallskip
	\begin{center}
		%		\resizebox{\columnwidth}{!}{%
			\begin{tabular}{  |c | c | c | c | c | c | c | c | c | c | c |}
				\hline  
				\multirow{2}{*}{\textbf{MAD}} & \multirow{2}{*}{\textbf{Method}} & 
				\multirow{2}{*}{\textbf{EER (\%)}} &
				\multicolumn{4}{c|}{\textbf{APCER(\%)@BPCER =}} & 
				\multicolumn{4}{c|}{\textbf{BPCER(\%)@APCER =}} \\
				\cline{4-11}
				& & & 1\% & 5\%  & 10\% & 20\% & 1\% & 5\% & 10\% & 20\%  \\ \hline
				\multirow{6}{*}{HRN~\cite{2022syn}} 
				& HGFM (w/o geometry) & 35.35 & 99.05 & 90.36 & 81.66 & 64.65 & 100.0 & 100.0 & 100.0 & 18.74 \\
				& HGFM (w/ ArcFace) & 35.53 & 99.05 & 89.98 & 82.23 & 64.84 & 100.0 & 100.0 & 100.0 & 46.23\\
				& HGFM (w/o local) & 35.53 & 98.86 & 89.79 & 82.41 & 63.71 & 100.0 & 100.0 & 100.0 & 19.28\\
				& HGFM (w/o blending) & 9.45 & 72.40 & 29.11 & 8.88 & 0.76 & 20.57 & 13.02 & 9.62 & 6.23\\
				& HGFM (w/ poisson) & 34.24 & 99.05 & 90.24 & 82.34 & 64.11 & 100.0 & 100.0 & 100.0 & \textbf{47.55}\\
				& HGFM & \textbf{35.54} & \textbf{99.43} & \textbf{90.37} & \textbf{82.42} & \textbf{65.03} & \textbf{100.0} & \textbf{100.0} & \textbf{100.0} & 46.04\\\hline   
			\end{tabular}%}\vspace{-4mm}
	\end{center}
\end{table*}

\begin{table}[ht]
	%	\vspace{-6.5mm}
	\caption{ Comparisons of FID and MMPMR@FMR=0.1\% values with different components on FERET dataset.}
	\vspace{-11pt}
	\label{ablation}
	\smallskip
	\begin{center}\resizebox{0.95\columnwidth}{!}{
			\begin{tabular}{ | c | c | c | c | c | c |}
				\hline
				\multirow{2}{*}{\textbf{Method}} & \multirow{2}{*}{\textbf{FID$\downarrow$}} & \multicolumn{4}{c |}{\textbf{MMPMR (\%)$\uparrow$}} \\ 
				\cline{3-6} & & \textbf{FN}~\cite{facenet} & \textbf{AF}~\cite{arcface} & \textbf{CF}~\cite{curricularface} & \textbf{EF}~\cite{elasticface} \\\hline
				HGFM (w/o geometry loss) & 40.7 & 58.9 & 84.1 & 92.4 & 88.6 \\\hline
				HGFM (w/ ArcFace) & 41.1 & 57.1 & 84.3 & 92.4 & 87.9 \\\hline
				HGFM (w/o local nets) & \textbf{38.1} & 50.5 & 77.7 & 88.1 & 81.1 \\\hline
				HGFM (w/o blending) & 50.6 & 58.4 & 68.4 & 92.6 & 86.2 \\\hline
				HGFM (w/ poisson) & 44.7 & 53.4 & 64.0 & 89.0 & 82.4 \\\hline
				HGFM & 40.5 & \textbf{59.9} & \textbf{85.8} & \textbf{93.4} & \textbf{88.9} \\\hline
		\end{tabular}}%\vspace{-3mm}
	\end{center}
\end{table}

\textbf{Visual Quality Assessments.} The visual appearance of morphed images is essential, which determines whether it can successfully deceive human observers. As shown in Table~\ref{mmpmr}, our method, though is not the best in terms of image quality, is comparable to the best counterpart methods. More importantly, it is significantly better than existing methods in terms of attack success rate. Hence, our method is superior in terms of the overall performance.
%As tabulated in Table~\ref{mmpmr} and illustrated in Fig.~\ref{fig:examples}, HGFM achieves higher PSNR and SSIM values and lower FID values. 
%shows the FID, SSIM, and PSNR values of HGFM compared with other state-of-the-art methods. Generally, lower FID values suggest higher image quality, indicating that morphs are more similar to contributors. Meanwhile, higher SSIM values suggest a greater structural similarity between morphed and contributing images, and higher PSNR values indicate lower image distortion. HGFM ranks high in visual quality assessment metrics (see Table~\ref{mmpmr}). 
As illustrated in Fig.~\ref{fig:examples}, landmark-based morphs exhibit noticeable artifacts around the hair and face, while GAN-based morphs have excessively smooth faces and lack texture details. While ReGenMorph~\cite{2021regenmorph} exhibits higher SSIM and PSNR values, it is noteworthy that its FID values are considerably higher compared to our method (lower FID values indicate better performance). And ReGenMorph shows blurred facial areas and ghost morphs. Conversely, the latest SOTA method MorDIFF~\cite{2023mordiff} morphs reproduce some facial details well, but there are strange artifacts in the hair area and red spots on the face of individual samples. This makes them susceptible to detection. In contrast, the morphed images generated by our method look much more realistic with less artifact. This is achieved by refining different facial regions separately using local morph networks and eliminating artifacts in areas outside the face with a mask-guided image blending module. In addition, the landmark morphed images provide pixel-level supervision and balance geometry and appearance information from both contributors. 

\begin{figure}[t]
	%	\vspace{-6mm}
	\centering
	\includegraphics[width=0.95\linewidth]{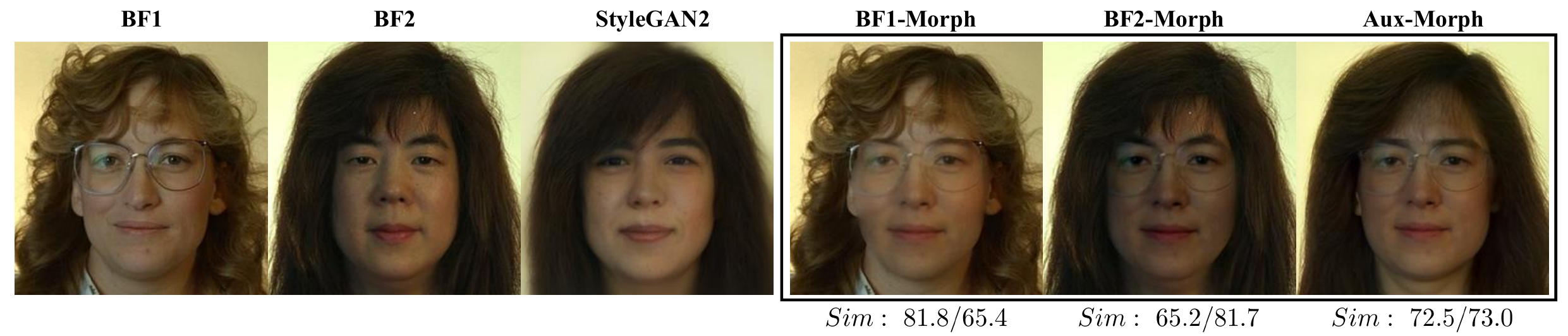}
	\caption{Comparison of different blending schemes. From left to right: The two contributing images (BF1 and BF2), the auxiliary image generated by StyleGAN2, the morphed images generated via image blending using BF1 and BF2 simultaneously, using BF1 and BF2 sequentially, and using the StyleGAN2-generated auxiliary image. As the similarity (\emph{Sim}) scores suggest, the last blending scheme achieves more balanced identity preservation between the two contributing subjects.}
	%		The morphed images within the black box are generated through image blending, sequentially employing contributing image 1 (BF1), contributing image 2 (BF2), and the auxiliary image generated by StyleGAN2 from left to right. The similarity scores of the morphed images and the two contributing images are calculated by ElasticFace~\cite{elasticface}. Observing the similarities, BF1-Morph is closer to contributing image 1, BF2-Morph aligns more with contributing image 2, whereas Aux-Morph exhibits a balanced similarity to both contributors.} 
\label{poisson}
\vspace{-5mm}
\end{figure}

%Additional comparison results with MorDIFF~\cite{2023mordiff} can be found in the supplementary materials.

\subsection{Different artifact removal strategies}
\begin{figure}[t]
	\centering
	\includegraphics[width=0.95\linewidth]{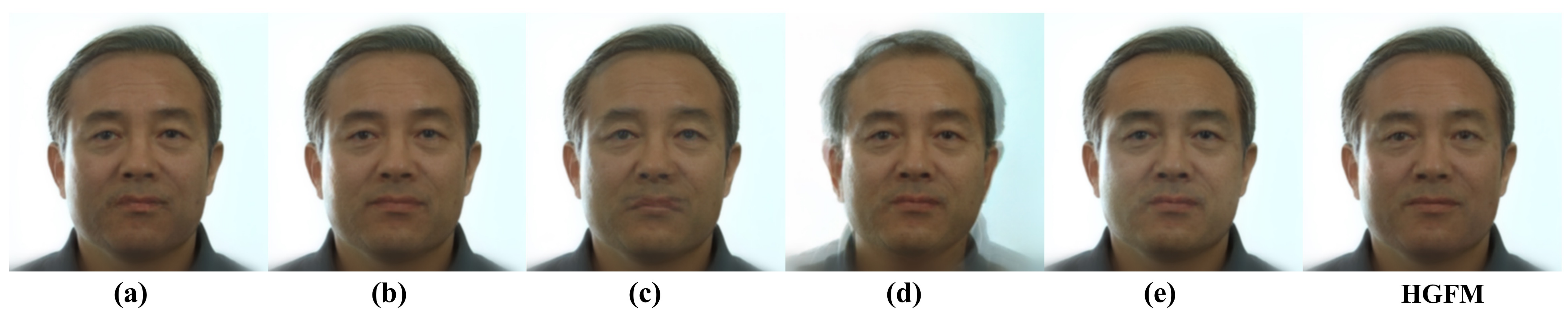}
	\caption{Morphed images generated by different variants of our HGFM method in the ablation study. (a) Without geometry loss, (b) Without combined identity loss, (c) Without local morph network and local loss, (d) Without mask-guided blending, and (e) Using Poisson blending instead. Texture degradation, poor geometry and lighting preservation, ghosting, blurring, and unnatural features can be observed in these results, respectively.}
	
	%		Visualization of the ablation study—analyzing the impact of each component in the HGFM: texture degradation without geometry loss (4. a), poor geometry and lighting preservation without combined identity loss (4. b), ghosting without local morph network and local loss (4. c), blurring from lack of mask-guided blending (4. d), unnatural features using Poisson blending (4. e).
	%		Qualitative results of ablation study. (a) without geometry loss, (b) standard identity loss (using ArcFace~\cite{arcface}), (c) without local nets and local loss (d) without mask-guided blending module and the mask loss, (e) replacement of mask-guided image blending with the Poisson image blending. Please zoom in to see the details.
	\label{exablation}
	\vspace{-5mm}
\end{figure}

To provide a better explanation for using StyleGAN2-generated images as auxiliary images for image blending in \textbf{Stage-II}, we use the Poisson image blending~\cite{Alpher12} method to simulate the mask-guided image blending module. We perform image blending using one of the contributing images. As depicted in  Fig.~\ref{poisson}, when directly using the contributing image 1 for background replacement, the resulting morphed image becomes more similar to contributing subject 1 and significantly less identical to contributing subject 2. This significantly reduces the performance of morphs and is inconsistent with the intention of the face morphing attacks. On the other hand, due to the varying size of the face region in the intermediate morphing image and the contributing images, a significant skin tone inconsistency occurs at the articulation of the face region and the neck using the Poisson image blending method (See Fig.~\ref{poisson} and Fig.~\ref{exablation} (e)). As mentioned above, instead of using the original contributing images, we employ StyleGAN2 morphs as background images. Moreover, the Poisson image blending method cannot simultaneously optimize the blended image and the generated intermediate morphed image. It also has limitations and drawbacks, relying on precise boundary information and incurring high computational overhead. In contrast, our HGFM introduces a mask-guided image blending module to mitigate the morphed image artifacts, ensuring the consistent facial region size between the intermediate morphed image and the background image.
%the mask-guided image blending module introduced in this paper effectively mitigates the artifacts in morphed images.

%This is the reason why this paper does not use the Poisson image blending method to solve the background artifacts.

\subsection{Ablation Study}
To investigate the effect of different modules of HGFM on morphing generation, we conduct an ablation study on FERET dataset. Specifically, we focus on four major components: $(i)$ the geometry loss, $(ii)$ the combined identity loss, $(iii)$ the local networks, and $(iv)$ the mask-guided image blending module. Table~\ref{ablation} shows the performance of the ablation study quantified using MMPMR and FID values. The qualitative results are shown in Fig.~\ref{exablation}. Quantitative results on MAD algorithms are given in Table~\ref{astudy}. According to the results, four conclusions can be summarized as follows. 

Firstly, using multiple pre-trained FR models to extract identity features enhances the identity preservation capability and the generalization ability of the morphed images against unknown FRSs. Secondly, supervision using an additional landmark-based morphed image is helpful to improve the visual quality of the morphed image along with the attack capability of FRSs. Thirdly, although there are still edge artifacts when combining local face patches using minimum pooling to resolve overlapping regions, the local networks effectively capture detailed information related to identity and refine local regions' facial texture. This significantly improves the attack performance. Fourthly, the mask-guided image blending module overcomes the issue of excessive color penetration from the intermediate morphed image into the auxiliary morphed image (background image), which can lead to loss of morphed content. It also produces a more natural and harmonious skin tone along the facial contour. As a result, artifacts are effectively removed, improving overall image quality.

\addtolength{\textheight}{-3cm}   % This command serves to balance the column lengths

\section{CONCLUSIONS}

This paper introduces HGFM, a new architecture for generating high-quality and high identity-preserving morphed images. The core of the proposed HGFM is the hierarchical generative network, which extracts detailed and consistency information of contributing images separately. HGFM combines the identity loss and geometry loss in the model optimization process, resulting in enhanced attack capability of morphed images against both known and unknown FRSs. 
%We use multiple pre-trained FR models to extract identity features with well-designed loss functions to constrain the morphing generation. 
HGFM also conceives a mask-guided image blending module to eliminate noticeable artifacts in areas outside the face to improve the visual quality further. To evaluate the attack potential of the proposed method, we have compared it with representative landmark-based and generation-based morphing methods to assess the visual quality of morphing images, identity preservation, and detection performance against the MAD algorithms. Quantitative and qualitative results show that HGFM performs better in preserving identity and visual quality than other methods. 

In future work, we will explore the refinement of the morphing network to extract global and local information more efficiently from both contributing images, and more effective strategies for combining the two types of information to generate higher-quality morphs.

%\section{AC}

%%%%%%%%%%%%%%%%%%%%%%%%%%%%%%%%%%%%%%%%%%%%%%%%%%%%%%%%%%%%%%%%%%%%%%%%%%%%%%%%
%\section{ACKNOWLEDGMENTS}

{
\bibliographystyle{ieee}

\bibliography{egbib}

\begin{thebibliography}{10}\itemsep=-1pt

\bibitem{2023diffusion}
Z.~Blasingame and C.~Liu.
\newblock Diffusion models for stronger face morphing attacks.
\newblock {\em arXiv preprint arXiv:2301.04218}, 2023.

\bibitem{borghi2021automated}
G.~Borghi, A.~Franco, G.~Graffieti, and D.~Maltoni.
\newblock Automated artifact retouching in morphed images with attention maps.
\newblock {\em IEEE Access}, 9:136561--136579, 2021.

\bibitem{elasticface}
F.~Boutros, N.~Damer, F.~Kirchbuchner, and A.~Kuijper.
\newblock Elasticface: Elastic margin loss for deep face recognition.
\newblock In {\em Proceedings of the IEEE/CVF Conference on Computer Vision and
  Pattern Recognition}, pages 1578--1587, 2022.

\bibitem{cao2022review}
S.~Cao, X.~Liu, X.~Mao, and Q.~Zou.
\newblock A review of human face forgery and forgery-detection technologies.
\newblock {\em Journal of Image and Graphics}, 27(4):1023--1038, 2022.

\bibitem{chen2017rethinking}
L.-C. Chen, G.~Papandreou, F.~Schroff, and H.~Adam.
\newblock Rethinking atrous convolution for semantic image segmentation.
\newblock {\em arXiv preprint arXiv:1706.05587}, 2017.

\bibitem{Alpher21}
R.~Chen, X.~Chen, B.~Ni, and Y.~Ge.
\newblock Simswap: An efficient framework for high fidelity face swapping.
\newblock In {\em Proceedings of the 28th ACM International Conference on
  Multimedia}, pages 2003--2011, 2020.

\bibitem{2019realistic}
N.~Damer, F.~Boutros, A.~M. Saladie, F.~Kirchbuchner, and A.~Kuijper.
\newblock Realistic dreams: Cascaded enhancement of gan-generated images with
  an example in face morphing attacks.
\newblock In {\em 2019 IEEE 10th International Conference on Biometrics Theory,
  Applications and Systems (BTAS)}, pages 1--10. IEEE, 2019.

\bibitem{2023mordiff}
N.~Damer, M.~Fang, P.~Siebke, J.~N. Kolf, M.~Huber, and F.~Boutros.
\newblock Mordiff: Recognition vulnerability and attack detectability of face
  morphing attacks created by diffusion autoencoders.
\newblock {\em arXiv preprint arXiv:2302.01843}, 2023.

\bibitem{2021regenmorph}
N.~Damer, K.~Raja, M.~S{\"u}{\ss}milch, S.~Venkatesh, F.~Boutros, M.~Fang,
  F.~Kirchbuchner, R.~Ramachandra, and A.~Kuijper.
\newblock Regenmorph: Visibly realistic gan generated face morphing attacks by
  attack re-generation.
\newblock In {\em Advances in Visual Computing: 16th International Symposium,
  ISVC 2021, Virtual Event, October 4-6, 2021, Proceedings, Part I}, pages
  251--264. Springer, 2021.

\bibitem{Alpher02}
N.~Damer, A.~M. Saladié, A.~Braun, and A.~Kuijper.
\newblock Morgan: Recognition vulnerability and attack detectability of face
  morphing attacks created by generative adversarial network.
\newblock In {\em 2018 IEEE 9th International Conference on Biometrics Theory,
  Applications and Systems (BTAS)}, pages 1--10, 2018.

\bibitem{frll}
L.~DeBruine and B.~Jones.
\newblock Face research lab london set, 2017.

\bibitem{arcface}
J.~Deng, J.~Guo, N.~Xue, and S.~Zafeiriou.
\newblock Arcface: Additive angular margin loss for deep face recognition.
\newblock In {\em Proceedings of the IEEE/CVF conference on computer vision and
  pattern recognition}, pages 4690--4699, 2019.

\bibitem{he2023optimal}
Q.~He, Z.~Deng, Z.~He, and Q.~Zhao.
\newblock Optimal-landmark-guided image blending for face morphing attacks.
\newblock In {\em 2023 IEEE International Joint Conference on Biometrics
  (IJCB)}, pages 1--9. IEEE, 2023.

\bibitem{curricularface}
Y.~Huang, Y.~Wang, Y.~Tai, X.~Liu, P.~Shen, S.~Li, J.~Li, and F.~Huang.
\newblock Curricularface: adaptive curriculum learning loss for deep face
  recognition.
\newblock In {\em proceedings of the IEEE/CVF conference on computer vision and
  pattern recognition}, pages 5901--5910, 2020.

\bibitem{2022syn}
M.~Huber, F.~Boutros, A.~T. Luu, K.~Raja, R.~Ramachandra, N.~Damer, P.~C. Neto,
  T.~Gon{\c{c}}alves, A.~F. Sequeira, J.~S. Cardoso, et~al.
\newblock Syn-mad 2022: Competition on face morphing attack detection based on
  privacy-aware synthetic training data.
\newblock In {\em 2022 IEEE International Joint Conference on Biometrics
  (IJCB)}, pages 1--10. IEEE, 2022.

\bibitem{Alpher15}
P.~Isola, J.-Y. Zhu, T.~Zhou, and A.~A. Efros.
\newblock Image-to-image translation with conditional adversarial networks.
\newblock In {\em Proceedings of the IEEE conference on computer vision and
  pattern recognition}, pages 1125--1134, 2017.

\bibitem{smdd}
M.~Ivanovska, A.~Kronov{\v{s}}ek, P.~Peer, V.~{\v{S}}truc, and B.~Batagelj.
\newblock Face morphing attack detection using privacy-aware training data.
\newblock {\em arXiv preprint arXiv:2207.00899}, 2022.

\bibitem{Alpher04}
T.~Karras, S.~Laine, and T.~Aila.
\newblock A style-based generator architecture for generative adversarial
  networks.
\newblock In {\em Proceedings of the IEEE/CVF conference on computer vision and
  pattern recognition}, pages 4401--4410, 2019.

\bibitem{Alpher14}
Z.~Ke, J.~Sun, K.~Li, Q.~Yan, and R.~W. Lau.
\newblock Modnet: Real-time trimap-free portrait matting via objective
  decomposition.
\newblock In {\em Proceedings of the AAAI Conference on Artificial
  Intelligence}, pages 1140--1147, 2022.

\bibitem{lucic2018gans}
M.~Lucic, K.~Kurach, M.~Michalski, S.~Gelly, and O.~Bousquet.
\newblock Are gans created equal? a large-scale study.
\newblock {\em Advances in neural information processing systems}, 31, 2018.

\bibitem{Alpher08}
S.~Mallick.
\newblock Opencv-based morph.
\newblock \url{https://learnopencv.com}, 2016.

\bibitem{matteo2014magic}
F.~Matteo, F.~Annalisa, and M.~Davide.
\newblock The magic passport.
\newblock In {\em IEEE International Joint Conference on Biometrics
  (IJCB’14)}, pages 1--7, 2014.

\bibitem{2021high}
M.~Pernu{\v{s}}, V.~{\v{S}}truc, and S.~Dobri{\v{s}}ek.
\newblock High resolution face editing with masked gan latent code
  optimization.
\newblock {\em arXiv preprint arXiv:2103.11135}, 2021.

\bibitem{frgc}
P.~J. Phillips, P.~J. Flynn, T.~Scruggs, K.~W. Bowyer, J.~Chang, K.~Hoffman,
  J.~Marques, J.~Min, and W.~Worek.
\newblock Overview of the face recognition grand challenge.
\newblock In {\em 2005 IEEE computer society conference on computer vision and
  pattern recognition (CVPR'05)}, volume~1, pages 947--954. IEEE, 2005.

\bibitem{feret}
P.~J. Phillips, H.~Wechsler, J.~Huang, and P.~J. Rauss.
\newblock The feret database and evaluation procedure for face-recognition
  algorithms.
\newblock {\em Image and vision computing}, 16(5):295--306, 1998.

\bibitem{Alpher12}
T.~Porter and T.~Duff.
\newblock Compositing digital images.
\newblock In {\em Proceedings of the 11th annual conference on Computer
  graphics and interactive techniques}, pages 253--259, 1984.

\bibitem{2020low}
L.~Qin, F.~Peng, S.~Venkatesh, R.~Ramachandra, M.~Long, and C.~Busch.
\newblock Low visual distortion and robust morphing attacks based on partial
  face image manipulation.
\newblock {\em IEEE Transactions on Biometrics, Behavior, and Identity
  Science}, 3(1):72--88, 2020.

\bibitem{Alpher09}
A.~Quek.
\newblock Facemorpher.
\newblock \url{http://www. facemorpher.com}, 2019.

\bibitem{raghavendra2017face}
R.~Raghavendra, K.~Raja, S.~Venkatesh, and C.~Busch.
\newblock Face morphing versus face averaging: Vulnerability and detection.
\newblock In {\em 2017 IEEE International Joint Conference on Biometrics
  (IJCB)}, pages 555--563. IEEE, 2017.

\bibitem{morphs}
E.~Sarkar, P.~Korshunov, L.~Colbois, and S.~Marcel.
\newblock Vulnerability analysis of face morphing attacks from landmarks and
  generative adversarial networks.
\newblock {\em arXiv preprint arXiv:2012.05344}, 2020.

\bibitem{2017biometric}
U.~Scherhag, A.~Nautsch, C.~Rathgeb, M.~Gomez-Barrero, R.~N. Veldhuis,
  L.~Spreeuwers, M.~Schils, D.~Maltoni, P.~Grother, S.~Marcel, et~al.
\newblock Biometric systems under morphing attacks: Assessment of morphing
  techniques and vulnerability reporting.
\newblock In {\em 2017 International Conference of the Biometrics Special
  Interest Group (BIOSIG)}, pages 1--7. IEEE, 2017.

\bibitem{facenet}
F.~Schroff, D.~Kalenichenko, and J.~Philbin.
\newblock Facenet: A unified embedding for face recognition and clustering.
\newblock In {\em Proceedings of the IEEE conference on computer vision and
  pattern recognition}, pages 815--823, 2015.

\bibitem{pytorchfid}
M.~Seitzer.
\newblock “pytorch-fid: Fid score for pytorch,” version 0.2.1.
\newblock \url{https://github.com/ mseitzer/pytorch-fid}, 2020.

\bibitem{venkatesh2021face}
S.~Venkatesh, R.~Ramachandra, K.~Raja, and C.~Busch.
\newblock Face morphing attack generation and detection: A comprehensive
  survey.
\newblock {\em IEEE transactions on technology and society}, 2(3):128--145,
  2021.

\bibitem{Alpher03}
S.~Venkatesh, H.~Zhang, R.~Ramachandra, K.~Raja, N.~Damer, and C.~Busch.
\newblock Can gan generated morphs threaten face recognition systems equally as
  landmark based morphs? -- vulnerability and detection, 2020.

\bibitem{Alpher16}
T.-C. Wang, M.-Y. Liu, J.-Y. Zhu, A.~Tao, J.~Kautz, and B.~Catanzaro.
\newblock High-resolution image synthesis and semantic manipulation with
  conditional gans.
\newblock In {\em Proceedings of the IEEE conference on computer vision and
  pattern recognition}, pages 8798--8807, 2018.

\bibitem{yi2019apdrawinggan}
R.~Yi, Y.-J. Liu, Y.-K. Lai, and P.~L. Rosin.
\newblock Apdrawinggan: Generating artistic portrait drawings from face photos
  with hierarchical gans.
\newblock In {\em Proceedings of the IEEE/CVF conference on computer vision and
  pattern recognition}, pages 10743--10752, 2019.

\bibitem{Alpher05}
H.~Zhang, S.~Venkatesh, R.~Ramachandra, K.~Raja, N.~Damer, and C.~Busch.
\newblock Mipgan—generating strong and high quality morphing attacks using
  identity prior driven gan.
\newblock {\em IEEE Transactions on Biometrics, Behavior, and Identity
  Science}, 3(3):365--383, 2021.

\bibitem{Alpher13}
K.~Zhang, Z.~Zhang, Z.~Li, and Y.~Qiao.
\newblock Joint face detection and alignment using multitask cascaded
  convolutional networks.
\newblock {\em IEEE signal processing letters}, 23(10):1499--1503, 2016.

\bibitem{zhang2023morphganformer}
N.~Zhang, X.~Liu, X.~Li, and G.-J. Qi.
\newblock Morphganformer: Transformer-based face morphing and de-morphing.
\newblock {\em arXiv preprint arXiv:2302.09404}, 2023.

\end{thebibliography}
}

\end{document}